\documentclass[conference]{IEEEtran}
\IEEEoverridecommandlockouts
\usepackage{cite}
\usepackage{amsmath,amssymb,amsfonts}
\usepackage{algorithmic}
\usepackage{graphicx}
\usepackage{textcomp}
\usepackage{subcaption}
\usepackage{float}
\usepackage{xcolor}
\usepackage{booktabs}
\usepackage{multirow}
\usepackage{makecell}
\usepackage{rotating}
\usepackage{lipsum} 
\usepackage{vcell}
\usepackage{caption} 

\usepackage{mdframed}
\mdfdefinestyle{MyFrame}{%
    linecolor=gray,
    innertopmargin=2pt,
    innerbottommargin=2pt,
    innerrightmargin=2pt,
    roundcorner=10pt,
    innerleftmargin=2pt,
    linewidth=1pt,
    backgroundcolor=white
}

\DeclareCaptionStyle{ruled}{labelfont=normalfont,labelsep=colon,strut=off}
\def\BibTeX{{\rm B\kern-.05em{\sc i\kern-.025em b}\kern-.08em
    T\kern-.1667em\lower.7ex\hbox{E}\kern-.125emX}}
\begin{document}
\setlength{\parskip}{0pt}
\title{On The Empirical Effectiveness of Unrealistic Adversarial Hardening Against Realistic Adversarial Attacks}

%

\author{\IEEEauthorblockN{ \qquad  1\textsuperscript{st} Salijona Dyrmishi \qquad }
\IEEEauthorblockA{\textit{University of Luxembourg} \\
salijona.dyrmishi@uni.lu}
\and
\IEEEauthorblockN{ \qquad  2\textsuperscript{nd} Salah Ghamizi \qquad }
\IEEEauthorblockA{\textit{University of Luxembourg} \\
salah.ghamizi@uni.lu}
\and
\IEEEauthorblockN{ \qquad 3\textsuperscript{rd}  Thibault Simonetto \qquad}
\IEEEauthorblockA{\textit{University of Luxembourg} \\
thibault.simonetto@uni.lu}
\and
\IEEEauthorblockN{\qquad \qquad \qquad \qquad \qquad \qquad \qquad  4\textsuperscript{th} Yves Le Traon \qquad }
\IEEEauthorblockA{ \qquad \qquad \qquad \qquad \qquad \qquad \textit{University of Luxembourg} \\
 \qquad \qquad \qquad \qquad \qquad \qquad yves.letraon@uni.lu}
\and
\IEEEauthorblockN{5\textsuperscript{th} Maxime Cordy \qquad }
\IEEEauthorblockA{\textit{University of Luxembourg} \\
maxime.cordy@uni.lu}
}

\maketitle

\begin{abstract}

While the literature on security attacks and defenses of Machine Learning (ML) systems mostly focuses on unrealistic adversarial examples, recent research has raised concern about the under-explored field of realistic adversarial attacks and their implications on the robustness of real-world systems. Our paper paves the way for a better understanding of adversarial robustness against realistic attacks and makes two major contributions. First, we conduct a study on three real-world use cases (text classification, botnet detection, malware detection) and seven datasets in order to evaluate whether unrealistic adversarial examples can be used to protect models against realistic examples. 
Our results reveal discrepancies across the use cases, where unrealistic examples can either be as effective as the realistic ones or may offer only limited improvement. Second, to explain these results, we analyze the latent representation of the adversarial examples generated with realistic and unrealistic attacks. We shed light on the patterns that discriminate which unrealistic examples can be used for effective hardening. We release our code, datasets and models to support future research in exploring how to reduce the gap between unrealistic and realistic adversarial attacks.

\end{abstract}


\section{{Introduction}}

Adversarial attacks are considered as one of the most critical security threats in Machine Learning (ML) \cite{xue2020machine, liu2018survey}. These attacks apply small perturbations to some original examples in order to produce \emph{adversarial examples}, specifically designed to fool the model decision. Adversarial attacks have been studied mostly in computer vision but also in various other domains like credit scoring\cite{ghamizi2020search},  cybersecurity \cite{aghakhani2020malware, pierazzi2020intriguing, sheatsley2020adversarial}, cyber-physical systems \cite{li2020conaml} and natural language processing~\cite{alzantot2018generating}. 

In order to enable the secure deployment of ML models in the real world, it is essential to properly assess their robustness to adversarial attacks and develop means to make models more robust. A common way to assess robustness is to empirically compute the model accuracy on the adversarial examples that an attack produced from a set of original examples. Similarly, the established way to harden ML models is adversarial hardening \cite{athalye2018obfuscated}, i.e. training processes that make models learn to make correct predictions on adversarial examples.


However, recent studies \cite{ghamizi2020search, tian2020exploring} have shown that in many domains, traditional adversarial attacks (e.g. PGD\cite{madry2017towards}) cannot be used for proper robustness assessment because these attacks produce examples that are not feasible (i.e. do not map to real-world objects). Indeed, while in computer vision the perturbations are simply independent pixel alterations that produce a similar image, in other domains the produced adversarial examples should satisfy specific domain constraints in order to represent real-world objects.


As a result, research has developed domain-specific adversarial attacks that either manipulate real objects through a series of problem-space transformations (i.e. \emph{problem-space attacks}) or generate feature perturbations that satisfy domain constraints (i.e. \emph{constrained feature space attacks}). These attacks produce examples that are realistic by design, however, at the cost of an increased computational cost compared to traditional attacks. This additional cost can be so high that it prevents the number of examples that ML engineers can use to assess and improve robustness.

In face of this dilemma between realism and computational cost, we pose the question whether we could improve model robustness against realistic examples through adversarial hardening on non-realistic examples. A positive answer would enable model hardening at affordable computational cost and without developing specific attacks that work in the particular subject domain. We study this question through an empirical study that spans over three domains where realistic attacks exist: \emph{text classification} where a problem-space attack (\emph{TextFooler} \cite{jin2020bert}) replaces words in a sentence with their synonyms; \emph{botnet traffic detection} where a constrained feature space attack (\emph{FENCE} \cite{chernikova2019fence}) applies realistic modifications to network traffic features; and \emph{malware classification} where a problem-space attack (\emph{AIMED} \cite{castro2019aimed}) modifies malware PE files. For each of these three domains, we generate examples using unrealistic attacks that are either domain-specific (e.g. DeepWordBug \cite{gao2018black} for text) or generic (e.g. PGD \cite{madry2017towards}). We, then, harden the corresponding ML model on these examples and assess the resulting robustness against realistic attacks.

Our results reveal that, in the botnet use case, adversarial hardening with unrealistic examples can effectively protect against realistic attacks. But in the other use cases, unrealistic examples bring limited to no benefit, regardless of the number of unrealistic examples used for hardening. To explain these apparent discrepancies, we analyze the properties of all produced examples, in particular their location in the embedded space, their direction, their aggressiveness and their perturbed features. Through our analysis, we reveal that unrealistic attacks can be   partially or completely useful if they  follow the same directions as realistic attacks and have the same aggressiveness levels, as in the text and botnet case. By contrast in the malware case, the unrealistic examples follow different directions and are considerably less aggressive compared to realistic examples, which explains their inability to protect against realistic attacks. 

To summarize, the contributions of this paper are:
\begin{itemize}
    \item Based on our  literature review  on realistic adversarial attacks (see Appendix \ref{sec:related}), covering both problem-space  and feature-space attacks, we select three use cases of realistic adversarial attacks and conduct the first study on adversarial hardening using unrealistic attacks that span over multiple domains.
    \item We reveal general insights that future research can explore to reduce the gap between unrealistic and realistic adversarial attacks, and support model hardening  against real-world adversarial threats.
    \item We provide a replication package including a collection of seven attacks, the seven datasets, and the implementation of our experimental protocol.\footnote{https://github.com/serval-uni-lu/realistic\_adversarial\_hardening} 
\end{itemize}

\section{{Problem Formulation}}
We consider a multi-class classifier $H: \mathcal X \longrightarrow \mathcal Y $ that maps an m-feature vector $x = \{x_1, x_2 ..., x_m\} \in \mathcal X$ to a class $y \in Y$.  
The function $h: \: \mathcal{X} \rightarrow \mathbb{R}$  predicts a probability distribution for $Y$.
The predicted class $H(x)$  can be derived as the class with the highest probability, such that  $H(x) = \underset {i \leq m}{argmax}\quad h_{i}(x)$. 

\subsection{{Attacks}}
The aim of an adversary is to change the predicted class $H(x)$ by the classifier (see Appendix \ref{app:threat-model} for the threat model). In the case of targeted attacks, the aim is to flip the prediction to a desired class that is predefined. In the case of untargeted attacks, it is sufficient that the label is different from its original prediction. For simplicity, we formalize below the case of targeted attacks only. Another distinction concerns the dimension the adversary works in.  We can classify all attacks into feature-space attacks (that alter input features) or problem space attacks (that directly manipulate domain objects). This refers to the case where the feature mapping between feature space and problem space is not differentiable and not invertible.

\subsubsection{Feature space attacks}
The attacker in feature space modifies directly the feature vector $x_0$ that is fed to the machine learning classifier $h(x)$. We distinguish between (traditional) unconstrained attacks and (domain-specific) constrained attacks.

\textbf{Unconstrained feature-space attacks}.
Given an initial feature vector $x_0$, the attacker aims to maximize the probability that the adversarial feature vector $\hat{x_0} = x_0 + \delta$ is classified as a predefined target  class  $t \neq H(x_0)$, where $\delta$ is a  perturbation vector. When performing an unconstrained attack, the adversary can perturb every feature as long as the generated adversarial example remains sufficiently close to the original example. The most popular metrics to measure similarity are $L_p$ norms that come from computer vision, however, in other domains specific metrics can also be defined (e.g. Cartella et al. \cite{cartella2021adversarial} define a metric for credit fraud detection that considers feature importance and human checks). We, then, define an unconstrained adversarial attack objective as: 
\begin{equation}
\begin{aligned}
\text{minimize} \quad D(x_0, \hat{x_0})\\
\text{such that}\quad  H(\hat{x_0}) = t
\end{aligned}
\end{equation}
where D is the distance metric of choice.

\textbf{Constrained feature-space attacks}. 
Many classification datasets are subject to constraints that come from inherent domain properties or from the way features are engineered. Different forms of constraint exist, including feature immutability (the adversary cannot alter specific features), data types, linear/non-linear relationships between multiple features, etc. These constraints pose additional restrictions for the adversarial examples to be considered valid (in addition to the perturbation size). Hence, constrained attacks perturb the original example in a direction that is valid with respect to the constraints. We denote the set of constraints by $\Omega$ and the set of feasible examples by $X_\Omega = \{ x \in X | x  \vDash \omega \in \Omega \}$. When $\Omega$ exactly translates all the constraints that exist in a domain, all examples in $X_\Omega$ may be produced in reality. Constrained adversarial attacks generate only examples in $X_\Omega$. Hence, the attack objective takes the form:

\begin{equation}
\begin{aligned}
\text{minimize} \quad D(x_0, \hat{x_0})\\
\text{such that}\quad  H(\hat{x_0}) = t \\
 \quad \hat{x_0} \in X_\Omega
\end{aligned}
\end{equation}

\subsubsection{Problem space attacks}
In this setting instead of dealing with numerical feature vectors, problem-space attacks directly manipulate physical (e.g. printed patches, sensors, sounds) or digital (e.g. pdf files, Android apps) domain objects. Let $Z$ be the set of domain objects. The feature mapping function $\phi: Z \longrightarrow X $ maps any $z \in Z$ to a feature representation $x \in X$. Problem space attacks aim to find a valid sequence of domain-object transformations $T  = T_n \circ T_{n-1} \circ ...\circ T_1$, such that $T(z)$ satisfies some problem space constraints $\gamma$. These constraints $\gamma$ encapsulate both general adversary concerns (e.g. similarity or preserved semantics) and domain-specific considerations (e.g. plausibility) \cite{pierazzi2020intriguing}. 
The problem-space attack objective is then:
\begin{equation}
\begin{aligned}
\text{minimize}  \quad { | T | \text{ where } T \in \mathcal{T}} \\
\text{such that:} \quad H(\phi(T(z))) = t\\
 \quad T(z) \vDash \gamma
\end{aligned}
\end{equation}
where $\mathcal{T}$ is the space of sequences of available transformations and $|T|$ is the length of a sequence $T$. 


\subsection{{Hardening methods}}
There are different approaches to harden a model against adversarial attacks. We focus on the methods that involve the inclusion of adversarial examples in the training process of the classifier $H(x)$ based on the original training dataset $D = \{(x_i, y_i)^N_{i=1}\}$ (See Appendix \ref{app:defender-model} for the defender's model). We name such methods \emph{adversarial hardening} and distinguish between three categories.

\textbf{Adversarial training} \cite{szegedy2013intriguing} is the most standard way to harden models. In application, it generates adversarial examples using an efficient method (e.g. FGSM \cite{goodfellow2014explaining}) and include these examples during the training process of the model, mixing the adversarial examples with the original data over the training epochs. The most established strategy \cite{madry2017towards} is to generate each time the worst-case adversarial examples (that achieve the highest loss) and, then, update the model learnable parameters to minimize the classification error over these adversarial examples. That is, adversarial training solves the following min-max optimization problem:
\begin{equation}
\begin{aligned}
min_{\theta} \quad \mathcal E(x,y)\sim D \left [ max_{\delta \in S}L(\theta, x + \delta, y) \right ]
\end{aligned}
\label{eq-4}
\end{equation}
where $(x, y) \sim D$ represents training data sampled from
the distribution $D$, $\delta \in S$ is the maximum allowed perturbation, $L$ is the model loss and $\Theta$ is the model hyperparameters. 


\textbf{Adversarial retraining} \cite{chen2020explore} is among the simplest techniques to defend against adversarial attacks. We use an attack $A$ over the training dataset $D$ to produce an adversarial training set $D^a = \{(\hat{x_i}, \hat{y_i})^N_{i=1}\}$ . 
We join the original and the adversarial training sets and retrain the classifier $H(x)$ over this augmented set $D^* = D \cup D^a$. It is to be noted that this is different from adversarial training as proposed by \cite{szegedy2013intriguing}, where new adversarial examples are generated and mixed with the original training set \emph{continuously}. Adversarial retraining is suited for both traditional machine learning algorithms and neural networks.



\textbf{Adversarial fine-tuning}\cite{jeddi2020simple} is a computationally efficient alternative to adversarial training. Indeed, while adversarial training resists well to even newly emerging attacks \cite{athalye2018obfuscated}, it is also very costly. On average, it needs 8–10 times more computational resources than the normal training of a neural network \cite{jeddi2020simple}. Therefore, some authors propose to perform adversarial fine-tuning, which is a mix of standard training and adversarial training. The idea of the adversarial fine-tuning is that we train for $e = e' + e''$ epochs: $e'$ epochs of standard training (using original training data) and, then, $e''$ epochs of adversarial training (using only adversarial examples).

\section{{Objectives and Research Questions}}

All the hardening methods mentioned earlier rely on the generation of adversarial examples in order to train models that correctly classify these examples. The application of these methods to protect models against real-world adversarial attacks would, ideally, necessitate the generation of realistic adversarial examples. Realistic adversarial methods -- be they constrained feature space attacks or problem space attacks -- are, however, computationally expensive and specifically designed for the subject domain and problem.  Table \ref{tab-generation-times} shows the time the attacks  in this study require generating a single adversarial example. Realistic attacks are 3.8 to 22650 times slower than unrealistic attacks. This largely inhibits the practical use of adversarial hardening methods, which generally require a large number of examples to protect the model effectively.

\begin{table}[ht!]
\centering
\caption{Adversarial examples' generation time for the attacks  considered in this study}
\label{tab-generation-times}
\begin{tabular}{@{}llll@{}}
\toprule
\textbf{Use-case}                          & \textbf{Attack}        & \textbf{Avg. time (s)} & \textbf{Relative time} \\ \midrule
\multirow{3}{*}{Text}       & DeepWordBug            & 0.15              & 1                      \\
                                           & TextFooler (Realistic) & 0.7               & 4.7                   \\
                                           & PWWS (Realistic)       & 0.57              & 3.8                    \\ \cmidrule(l){2-4} 
\multirow{2}{*}{Botnet}  & PGD                    & 0.12              & 1                      \\
                                           & FENCE (Realistic)      & 0.95              & 7.9                    \\ \cmidrule(l){2-4} 
\multirow{3}{*}{Malware} & PGD                    & 0.002             & 1                      \\
                                           & MoEvA2                 & 36                & 18000                  \\
                                           & AIMED (Realistic)      & 45.3              & 22650                  \\ \bottomrule
\end{tabular}
\end{table}

In addition to runtime costs, one should consider the engineering effort behind the generation of adversarial examples.  We identify two components in this case: the design cost to create new attacks and the engineering costs to apply them.  The cost of designing new attacks is closely related to the level of realism expected as an outcome. The mapping from problem space to feature space is generally not invertible, hence many realistic attacks need to work in problem space.  As Pierazzi et al. \cite{pierazzi2020intriguing} point out, such problem-space attacks necessitate the sequential application of problem-space (domain-specific) transformations that alter original problem-space objects into valid objects that the model misclassifies. To be realistic, these transformations must satisfy a series of conditions including: being applicable by the attacker, preserving (to some extent) the semantics of original objects, being plausible (robust to human analysis) and being robust to preprocessing. These conditions together make the engineering of the transformations challenging. Furthermore, due to the not invertible mapping, the transformations can only be applied in black-box and do not benefit from internal model feedback to drive the transformations. Ultimately, the development and assessment of problem-space transformations can only be achieved through intense experimentations. This is why the literature has primarily focused on generic, unconstrained feature-space attacks and why problem-space attacks are (although rarely) developed in specialized fields of research. Developing realistic attacks when they do not already exist is a cumbersome process. 

Once these attacks are designed, they differ as well on the engineering to apply them. Some attacks like PGD are already part of standard libraries. In this case, the only  costs foreseen are related to data preprocessing. Generic constrained  attacks like MoEvA2 require their users to write constraints in analytical form, which typically requires the involvement of a domain expert. And lastly, there are problem-space attacks for critical domains that require particular security precautions. As an example, attacks against malware detectors  have to be operated within an isolated sandbox (virtual machine) because the attacker risks corrupting its own system.

\subsection{{Research Questions}}

Our research investigates whether methods that generate unrealistic examples -- and do so at lesser computational cost -- can be used to harden models against realistic attacks. 
\begin{description}
\item[\bf RQ1:] \em Can unrealistic adversarial hardening improve model robustness to realistic attacks?
\end{description}
We study this question across three use cases (cf. Sections \ref{sec:text}, \ref{sec:botnet}, and \ref{sec:malware}) that span over three domains: text classification, botnet traffic detection, and malware detection. In each of these use cases, we consider a realistic adversarial attack (a problem space attack for text and malware, and a constrained feature space attack for botnet) and at least one unrealistic attack. We apply adversarial hardening using the unrealistic attack(s) and compute the prediction accuracy of the resulting model against adversarial examples generated by the realistic attack. We compare this robust accuracy against the robust accuracy of the model hardened through realistic generation methods. In each of our experiments, we allow each method (both realistic and unrealistic) to generate the same number of adversarial examples. A gap between the two accuracy values would indicate that unrealistic methods cannot harden models as much as realistic methods enable it.

We, next, investigate whether increasing the computational budget of unrealistic model hardening (measured as the number of examples generated) can  improve robust accuracy. Indeed, while unrealistic methods are computationally cheaper than realistic methods, the generation of additional examples might further help model hardening.
\begin{description}
\item[\bf RQ2:] \em Does generating more unrealistic adversarial examples help further hardening models against realistic attacks?
\end{description}
To answer this question, we study the trend of robust accuracy when we increase the number of adversarial examples used for hardening. A monotonic trend would motivate research on increasing the efficiency of unrealistic adversarial attacks, whereas an asymptotic behavior would indicate that unrealistic methods may never reach the effectiveness of realistic methods.

After observing the robust accuracy, our third and last research question investigates the difference between the realistic and unrealistic examples that explain the obtained results.
\begin{description}
\item[\bf RQ3:] \em What properties of the generated examples affect the hardening results?
\end{description}

We study, more precisely, where the examples generated by realistic and unrealistic attacks are located compared to clean examples and to each other. We do this initially through t-SNE embeddings as a means of visualizations. We expand further our study to go beyond visualizations and combine t-SNE with quantitative metrics like cosine similarity  and aggressiveness (defined in Section \ref{sub-sec:agg}). We use these quantitative metrics to highlight any insights related to the position of adversarial examples. We conclude by analyzing the distribution of changed features for each feature-space attack (Malware and Botnet use cases).

\subsection{Metrics used}
\label{sub-sec:agg}

\textbf{Clean and Robust accuracy.} We use the term clean accuracy to refer to the standard test accuracy. We use the term robust accuracy to refer to the ratio between the number of correctly predicted adversarial examples over the total number of adversarial examples.

\textbf{t-SNE visualization.} We first reduce data dimensions to 30 with
PCA, considering that t-SNE with high dimensional data is computationally expensive. Later we use the PCA output to generate 2-dimensional data with t-SNE and plot the results. We vary the perplexity values (10,20,30,40,50) and run the experiments with 5 different random states. One  plot for each use case is shown in the paper, and a comment on how the hyperparameters affected the results are added in the respective sections.

\textbf{Intra-cluster dispersion scores.}
We calculate the intra-cluster dispersion score as the sum of distances from adversarial examples to the centroid of adversarial examples generated by the  same attack.

\textbf{Cosine similarity} measures the similarity between two vectors' directions. For each example, we compute the cosine similarity between the vectors formed by the application of a realistic and an unrealistic attack.  We, then, average the cosine similarity over all original examples. Since cosine similarity always falls between -1 and 1, we can compare this metric across the different use cases.

\textbf{Aggressiveness} is the metric we introduce in order to compare how aggressively each attack works, i.e. how far the attack creates adversarial examples from the original inputs (this is the opposite notion of  attack confidence\cite{pierazzi2020intriguing} or adversarial friendliness \cite{zhang2020attacks} that previous research has introduced). Let $X^{*} \subset X$ such that $x \in X^{*}$ if $y=t$ and $H(x)=t$. Then  $x^{nn} = \underset {x^{*} \in X^{*} }{argmin}\quad D(x, x^{*})$. We can now define the $``aggressiveness"$ as in Equation \ref{equation-metric}. 

\begin{equation}
\label{equation-metric}
\begin{aligned}
aggressiveness = \dfrac{D(x, \hat{x})}{D(x,x^{nn})}
\end{aligned}
\end{equation}

Traditional distance metrics like Euclidean distance are unfortunately not uniformized over their data space, which makes it difficult to compare these distances across multiple use cases. Therefore, we divide the Euclidean distance between the initial example ($x$) to its adversarial counterpart ($\hat{x}$) by the distance between $x$ and the closest original example ($x^{nn}$) that is predicted correctly in the target class of the attack. The distance to $x^{nn}$ serves as a reference point for the minimum known distance to cross the decision boundary.  At the same time, it allows us to have a standardized metric that can be applied to three use cases and compare between them.
Having defined this metric, we compute, for each attack, the mean aggressiveness over all original examples.

We calculate the above metrics using the feature space representation of the examples (botnet, malware) and when this is not possible (text), we use the sentence embeddings calculated from the models' last hidden layer.

\subsection{{Use Case Selection}}
\label{ref:cases}

To conduct our experiments, we selected application domains and learning tasks where: (1) inputs are inherently constrained, (2) open-source datasets are available (3) realistic attacks have been proposed with available implementations. We identified many domains with datasets and constraints that originate  from the problem space itself (and these constraints also translate into the feature space) and from the way features are engineered. For instance, in the botnet use case, the feature engineering results in features (a) the number of bytes, (b) the number of connections, and (c) the number of bytes per connection. All inputs must therefore satisfy the new constraints a = b * c to be realistic.
This identification phase yielded a set of papers that we present in Appendix \ref{sec:related}. However, we have observed three concerns that narrowed down our choice:
\begin{itemize}
    \item There are only a few domains where realistic attacks (either in the problem space or the feature space) have been designed, and even fewer with a publicly available implementation. For some domains, problem space or constrained feature space attacks exist but are not entirely realistic because they produce partially valid domain objects or consider only a subset of all constraints that exist. 
    \item An alternative to using previously proposed attacks is to tailor a generic constrained attack framework (e.g. \cite{chernikova2019fence, ghamizi2020search}) with constraints specific to the target domain. This requires, however, either detailed documentation of the dataset features (which often overlooks the definition of constraints) or rare domain expertise that we may not have access to.
    \item Even when they exist, realistic attacks can have a high computational cost because they either have to resolve constraints or perform complex transformations on domain objects. Therefore, when we have several alternatives for a given domain, we tend to choose the fastest realistic attack.
\end{itemize}
Our selection process resulted in three pairs of domains/tasks and attacks that we analyze hereafter. The first pair is \emph{text classification} and the \emph{TextFooler} attack. This problem-space attack replaces words with synonyms and makes sure that sentences remain syntactically and semantically consistent. The second is \emph{botnet traffic detection} and the \emph{FENCE} attack. FENCE is a constrained feature space attack that  generates realistic features of network traffic flows. The third is \emph{Windows malware detection} and the \emph{AIMED} attack. AIMED is a problem-space attack based on a genetic algorithm that alters the PE file of a malware without changing the malicious behavior.

For each of these use cases, we conduct our experiments using state-of-the-art models and established datasets. We, moreover, rely on the model hardening processes that have originally been used to counter the realistic attacks we consider.

\section{{Text Classification}}
\label{sec:text}
The first use case that we consider is natural language processing, in particular text classification. 
Morris et al. \cite{morris2020reevaluating} define four types of constraints for a successful and realistic adversarial example in natural language tasks: \textit{ semantics, grammaticality, overlap, } and \textit{non-suspicion} (see Appendix \ref{app:text-constraints}).   
Hence, we consider as realistic any problem space attack that satisfies Morris et al.'s four types of constraints, and as unrealistic any attack that does not satisfy any of those constraints. In this use case, feature space attacks cannot guarantee the realism of the produced examples, as the transformations from text to feature space are generally not invertible.



\subsection{{Experimental Protocol}}


\textbf{Realistic adversarial attacks (to protect against).} Attacks based on synonym replacement respect all four constraints described above \cite{morris2020reevaluating} and are, therefore, considered realistic. We use two such attacks: TextFooler \cite{jin2020bert}, and PWWS \cite{ren2019generating}. TextFooler ranks the words based on their importance and replaces top-ranked words with semantically similar words. For example, from the sentence ``\textit{poorly executed comedy}'' TextFooler could produce the adversarial sentence ``\textit{\textbf{faintly} executed comedy}''. PWWS uses word saliency and prediction probability to decide the word to be substituted and the replacement order for the synonyms.

\textbf{Unrealistic adversarial attack (for hardening).} We use the problem-space attack named DeepWordBug \cite{gao2018black}, which is based on character substitution. DeepWordBug selects the words with the highest influence on the model's decision, then applies to them character substitution transformations. For example, from the sentence ``\textit{poorly executed comedy}'' DeepWordBug could produce the adversarial sentence ``\textit{p\textbf{K}orly executed comedy}''. Character-level substitution obviously does not satisfy all four realism constraints and, therefore, DeepWordBug adversarial examples are unrealistic.

We evaluated the realism of the DeepWordBug and TextFooler adversarial examples by correcting their spelling with the LanguageTools proofreading service. This step filtered 87.4\% of the successful  DeepWordBug adversarial examples,  compared to only 7.31\% for TextFooler. This confirms our claim that DeepWordBug examples are unrealistic and easily detectable by language checkers.

\textbf{{Datasets and Model.}} We use Rotten Tomatoes, Tweet Offensive and AG News datasets to train for 5 epochs a DistilBert model. You can find more details about the datasets and models in Appendix \ref{app:text_exp}.

\textbf{{Hardening strategy.}}
For hardening the models, we use one clean epoch at the beginning of the training process, followed up with four epochs where the training samples are the adversarial examples generated by either DeepWordBug or TextFooler. Hence, this is a case of adversarial fine-tuning with five epochs (1 clean epoch + 4 adversarial).  We use TextAttack v0.2.15 library by \cite{morris2020textattack} for our experiments and keep their default parameters for DeepWordBug and TextFooler when performing fine-tuning. 

\textbf{{Evaluation.}}
For each fine-tuned model, we evaluate the clean and robust accuracy against  TextFooler and PWWS. Again, we use the default hyperparameters from TextAttack library for the evaluation.
For validation set, we use the default split provided by TextAttack for Rotten Tomatoes (1066 samples) and Tweet Offensive (1324 samples). For AG News, we select randomly 1000 samples. 



\subsection{{Results -- RQ1 (Hardening)}}

In Table \ref{tab-text}, we compare the robust accuracy of the models against the realistic attacks TextFooler and PWWS, of the clean model, the model fine-tuned with DeepWordBug, and the model fine-tuned with TextFoolder.

We, first, observe that, although DeepWordBug manages to increase robust accuracy in all cases compared to the clean model, it does so with significantly less effectiveness than TextFooler in all cases. For example, DeepWordBug increases the robust accuracy on PWWS adversarial sentences by +7.35\% (from 12.22\% to 19.57\%) for Rotten Tomatoes,  +9.56\% for Tweet Offensive, and +7.5\% for AG News. By contrast, fine-tuning with TextFooler improves the robustness against PWWS by 15.29\%, 26.5\%, and 13.35\%, respectively. Similar observations are made when TextFooler is the attack to protect against.

Interestingly, TextFooler manages to harden the model against PWWS as well as it does against TextFooler itself. This indicates that, in this text classification use case, the protection brought by a given realistic attack generalizes to other realistic attacks. Though unrealistic attacks like DeepWordBug have a similar effect across the two realistic attacks, their benefits are far from those obtained with a realistic attack.

Note that fine-tuning with any of the two attacks has a negligible impact on the clean performance of the model, indicating that adversarial hardening, in this case, does not hurt performance against benign sentences. 





\begin{table}[ht!]
\centering
\caption{Clean and robust accuracy (in \%) of the clean model and the adversarially fine-tuned models, over three datasets. The first column of results represents the clean accuracy, and the remaining columns represent the robust accuracy against TextFooler and PWWS.}
\label{tab-text}
\begin{tabular}{ll|lll} 
\toprule
Dataset                          &                 & \multicolumn{2}{c}{Attacked by} &                 \\ 
\hline
                                 & Trained against & -     & TextFooler              & PWWS            \\ 
\midrule
\multirow{3}{*}{Rotten tomatoes} & Clean           & 83.68 & 5.49                    & 12.22           \\
                                 & DeepWordBug     & 83.4  & 9.34                    & 19.57           \\
                                 & TextFooler      & 80.49 & \textbf{17.72}          & \textbf{27.51}  \\ 
\cline{2-5}
\multirow{3}{*}{Tweet Offensive} & Clean           & 81.04 & 11.56                   & 18.45           \\
                                 & DeepWordBug     & 77.11 & 19.2                    & 28.01           \\
                                 & TextFooler     & 78.47 & \textbf{43.21}          & \textbf{44.95}  \\ 
\cline{2-5}
\multirow{3}{*}{AG News}         & Clean           & 89.6  & 16.07                   & 32.37           \\
                                 & DeepWordBug     & 89.8  & 17.29                   & 39.87           \\
                                 & TextFooler     & 89.9  & \textbf{26.14}          & \textbf{45.72}  \\
\bottomrule
\end{tabular}
\end{table}


\subsection{{Results -- RQ2 (Budget)}}

We investigate if generating more adversarial examples can help DeepWordBug achieve the same hardening effectiveness as TextFooler. We repeat our experiments with DeepWordBug for up to 20 epochs (instead of 5 initially), and we record for each epoch the robust accuracy against adversarial examples generated by TextFooler and PWWS. The increased number of epochs from 5 to 20 means we are multiplying the number of generated examples by 4, as we generate one example per original input in each epoch. 

\begin{figure*}
  \begin{subfigure}{0.33\linewidth}
  \includegraphics[width=\linewidth]{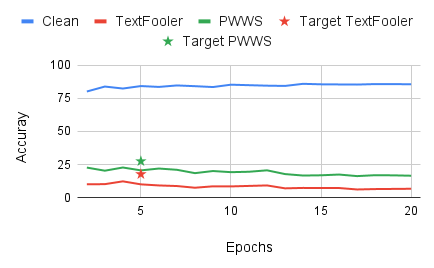}
    \caption{Rotten Tomatoes}
  \end{subfigure}\hfill
  \begin{subfigure}{0.33\linewidth}
  \includegraphics[width=\linewidth]{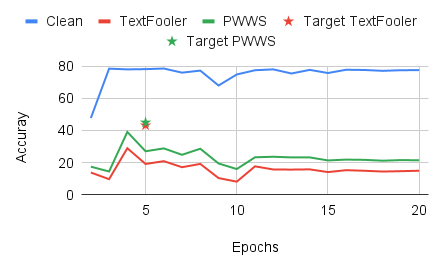}
    \caption{Tweet Offensive}
  \end{subfigure}\hfill
    \begin{subfigure}{0.33\linewidth}
  \includegraphics[width=\linewidth]{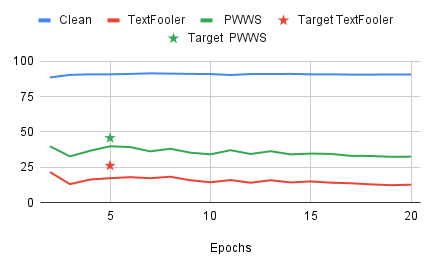}
    \caption{AG News}
  \end{subfigure}
  \caption{Robust accuracy of models adversarially fine-tuned with DeepWordBug over 20 epochs for the three datasets. Red line is robust accuracy against TextFooler, green line is against PWWS, blue line is clean accuracy. Stars represent the robust accuracy achieved by a 5-epoch TextFooler-based fine-tuning against TextFooler (red star) and PWWS (green star).}
  \label{fig-text-rq2}
  \end{figure*}

We show the results for the three datasets in Figure \ref{fig-text-rq2}. Though the robust accuracy values do fluctuate between the first and 15th epochs, the robustness that DeepWordBug-based hardening achieves never reaches the same level as the robustness previously achieved by hardening with TextFooler. Compared to our initial number of 5 epochs, increasing the number of epochs does not bring positive effects. This might indicate that the unrealistic attack generates examples confined to a limited space, and that a significant number of realistic adversarial examples lie outside this space. We investigate this further in RQ3.
 


\subsection{Results -- RQ3 (Properties)}

Following our results above, we hypothesize that the adversarial examples generated by TextFooler target different areas that the DeepWordBug adversarial examples do not reach. To facilitate the investigation of this hypothesis, we focus on Rotten Tomatoes and Tweet Offensive as these datasets are binary (AG News is multi-class). We analyze the sentence embeddings obtained from the last layer latent representation of the clean model. 



\paragraph{\textit{Embeddings of the adversarial examples}}

\begin{figure}[ht!]
\centering
  \includegraphics[width=\linewidth]{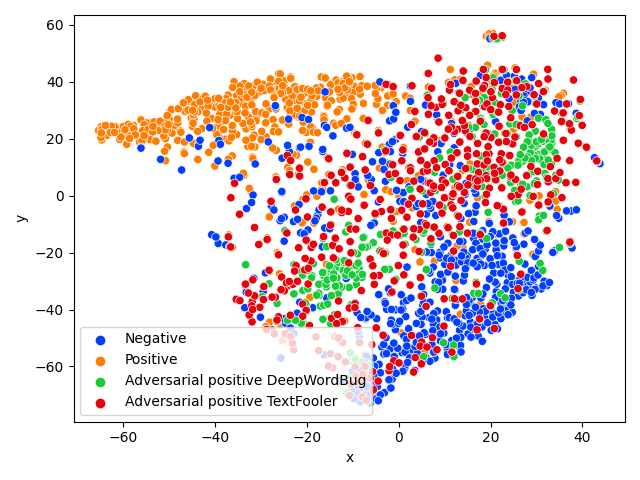}
  \caption{t-SNE visualization of original examples and the successful adversarial examples generated by DeepWordBug and TextFooler for Rotten Tomatoes positive reviews.}
  \label{fig-text-rq3}
  \end{figure}

To compare the embeddings of the different clean and adversarial examples, in Figure \ref{fig-text-rq3}, we visualize the t-SNE embeddings for the positive  reviews  in Rotten Tomatoes. Varying the value of the perplexity and using four other random states did not impact majorly the general formed structures. The hyperparameters affected only the graph stretching or graph rotations. We see that all adversarial examples generally lie between the two main class clusters (orange and blue dots), indicating that the attacks indeed navigate around the decision boundary. However, the DeepWordBug examples are closely grouped together, with a dense concentration at specific areas. Meanwhile, TextFooler examples have a wider spread throughout the decision boundary. The rest of the plots for  Rotten Tomatoes and Tweet Offensive are shown in Appendix \ref{app:plots} and corroborate our findings from Rotten Tomatoes.


This qualitative analysis confirms our hypothesis that TextFooler examples cover a larger portion of the space that the DeepWordBug examples never reach. This is the reason why DeepWordBug offers only limited improvement against realistic attacks.

\paragraph{\textit{Clustering analysis of the adversarial examples}}

To complement our analysis, we present in Table \ref{tab-text-cluster2} the intra-cluster dispersion score of the adversarial examples generated by DeepWordBug and TextFooler before applying the dimension reduction with t-SNE. 
A higher intra-cluster dispersion score indicates that the cluster is more spread over the latent space. This quantitative analysis confirms that adversarial examples generated by TextFooler cover more of the latent space. For instance, on the Rotten Tomatoes dataset, the TextFooler adversarial examples produced from positive original examples have an intra-cluster dispersion score 71.7\% higher than the DeepWordBug adversarial examples produced from the same original examples. 

\begin{table}[ht!]
\centering
\caption{Intra-cluster dispersion scores of adversarial examples generated by DeepWordBug and TextFooler for each class. A higher score indicates that the cluster is more spread.}
\label{tab-text-cluster2}
\begin{tabular}{llll} 
\toprule
Dataset                          & Class         & DeepWordBug & TextFooler     \\ 
\midrule
\multirow{2}{*}{Rotten Tomatoes}    & Negative   & 173899       & 241840        \\
                                 & Positive      & 155892       & 267629        \\
\midrule
\multirow{2}{*}{Tweet Offensive} & Offensive     & 221589       & 452303        \\
                                 & Non-offensive & 379399       & 401270        \\
\bottomrule
\end{tabular}
\end{table}

\paragraph{\textit{Perturbation size and direction}}

Table \ref{cosine-text} gives the mean and standard deviation values for cosine similarity and aggressiveness of adversarial examples. For both datasets, DeepWordBug and TextFooler generate adversarial examples that follow very similar directions, considering that their cosine similarities have high values, respectively 0.89 for Rotten Tomatoes  and 0.92 for Tweet Offensive. Additionally, they are similarly aggressive, with their mean aggressiveness score differing only by 0.06 for Rotten Tomatoes and 0.02 for Tweet Offensive.  This explains why DeepWordBug is still able to partially robustify the models that are faced with TextFooler examples.

\begin{table}[ht!]
\centering
\caption{Mean and standard deviation values for cosine similarity and aggressiveness of adversarial examples. }
\label{cosine-text}
\begin{tabular}{llllll}
\hline
\textbf{Dataset} & \textbf{Attack} & \multicolumn{2}{l}{\textbf{Cosine similarity}} & \multicolumn{2}{l}{\textbf{Aggressiveness}} \\
                 &                 & \multicolumn{2}{l}{\textbf{with TextFooler}}   &                      &                      \\ \hline
                 &                 & Mean        & \multicolumn{1}{l|}{Std}         & Mean                 & Std                  \\
Rotten           & DeepWordBug     & 0.89        & \multicolumn{1}{l|}{0.10}        & 0.90                 & 0.24                 \\
Tomatoes         & TextFooler      & 1           & \multicolumn{1}{l|}{0}           & 0.84                 & 0.27                 \\ \hline
Tweet            & DeepWordBug     & 0.92        & \multicolumn{1}{l|}{0.09}        & 0.78                 & 0.34                 \\
Offensive        & TextFooler      & 1           & \multicolumn{1}{l|}{0}           & 0.80                 & 0.37                 \\ \hline
\end{tabular}
\end{table}

\subsection{{Conclusion for Text Classification}}



Our evaluation confirms that unrealistic adversarial hardening does not suffice to protect against realistic adversarial attacks. Unrealistic adversarial hardening leads to models up to 16.94\% less robust than realistic adversarial hardening against real attacks. We also demonstrate that using more unrealistic adversarial examples in the hardening (up to four times more) does not lead to any further improvement of the robustness of the models against realistic attacks. We justify that we cannot fully replace realistic adversarial examples with unrealistic because unrealistic adversarial examples do not cover the feature space as well as the realistic examples. We demonstrate how realistic examples are widely spread, less clustered and better cover both classes of the binary classification task than unrealistic adversarial examples.  However, the unrealistic examples have similar direction and aggressiveness level as realistic examples, hence they still manage to provide some level of robustification. 

\section{{Botnet Traffic Detection}}
\label{sec:botnet}
The detection of botnet traffic is a successful application case of machine learning \cite{livadas2006usilng,tuan2020performance}. The resulting detectors typically learn from the network flows between the source and destination IP addresses that botnet and normal traffic generate. Network traffic is naturally subject to validity constraints. For example, the TCP and UDP packet sizes have a minimum byte size, and a change in the number of packets sent always brings a change in the total bytes sent. We are unaware of any problem space attacks for botnet detection. We, therefore, consider a constrained feature space, FENCE \cite{chernikova2019fence}, that claims to be realistic. The FENCE framework guarantees that it produces adversarial examples in feasible regions of the feature space, therefore they can be projected back to the raw input space. The projection operator needs to be defined for each application domain separately.

\subsection{{Experimental Protocol}}

\textbf{Realistic adversarial attack (to protect against).} We use the FENCE attack designed by \cite{chernikova2019fence}. FENCE is initially a generic framework that has been tailored to botnet traffic detection, and we reuse the original implementation presented in \cite{chernikova2019fence}. FENCE is a gradient-based attack that includes modifications such as mathematical constraints and projection operators to create feasible adversarial examples. 
As denoted in the original paper, a family of features is a set of features related by one or more constraints.
The algorithm starts by finding the feature of maximum gradient, referred as the representative feature, and its respective family.
The representative feature  of the family is updated with a value $\delta$, and the other members are updated based on $\delta$ and the constraints definitions. The value $\delta$ is found through a binary search such that the global perturbation does not exceed the maximum distance threshold.
Therefore, the constraints are satisfied by construction.


\textbf{Unrealistic adversarial attack (for hardening).} We use the PGD implementation from ART \cite{nicolae2018adversarial}. PGD is a traditional iterative gradient attack that adds a perturbation based on gradient direction at every iteration. We modify the implementation from ART to stop PGD updates at the moment an adversarial example is found. This modification is equivalent to Friendly Adversarial Training presented by \cite{zhang2020attacks} which does not degrade the clean accuracy of the models. 

As a preliminary check, we have evaluated the attacks’ level of realism using the  analytical feature constraints derived from Simonetto et al.  for botnet datasets. Respecting these constraints is a necessary but not sufficient condition for having realistic examples, since the problem space is more restricted than the feature space. Failure to satisfy  them indicates unrealistic examples. None of PGD examples respect the constraints and 100\% of FENCE examples respect the constraints. In addition, by design, it is always possible to project FENCE adversarial examples to their raw input space. This confirms that PGD attack is unrealistic and FENCE attack is realistic.

\textbf{{Datasets and Model.}} We use Neris, Rbot and Virut from standard CTU-13 dataset to train a neural network architecture with dense layers from the original FENCE paper \cite{chernikova2019fence} for 10 epochs. You can find more about the datasets in Appendix \ref{app:botnet_exp}.

\textbf{{Hardening strategy.}}
We harden the model performing adversarial training for 10 epochs. For both attacks, we use 5 iterations to generate the adversarial examples and a perturbation budget of $\epsilon=12$ under a $L_2$ norm. The $L_2$ distance is the one from the original FENCE paper \cite{chernikova2019fence}. 

\textbf{{Evaluation.}}
For evaluation, we use 100 iterations for both PGD and FENCE. In the botnet domain, the adversary is interested on passing a flow as normal traffic and not vice-versa. Therefore, we apply adversarial attacks that target the benign botnet class.

\subsection{{{Results -- RQ1 (Hardening)}}}

Table \ref{tab-botnet-rq1} compares the robust accuracy (when attacked by FENCE) of the clean models, the models adversarially trained with PGD, and the models adversarially trained with FENCE.  We observe that the PGD-trained model is perfectly robust against the FENCE attack. Across the three datasets, the robust accuracy jumps to 100\% after adversarial training with PGD (just like with FENCE). Meanwhile, the clean performance of the models is not affected. This reveals that even the unrealistic, unconstrained attack can protect the model as effectively as the realistic attack.


\begin{table}
\centering
\caption{Clean performance (measured with ROC AUC) and robust accuracy (against FENCE applied on botnet examples) of the original botnet detection model and the adversarially hardened models.}
\label{tab-botnet-rq1}
\begin{tabular}{ll|ll} 
\toprule
\multicolumn{1}{c}{\multirow{2}{*}{\begin{tabular}[c]{@{}c@{}}\textbf{Training }\\\textbf{ scenarios}\end{tabular}}} & \textbf{}      & \multicolumn{1}{c}{} & \textbf{\textbf{Attacked by}}  \\
\multicolumn{1}{c}{}                                                                                                 & \textbf{}      & \textbf{Roc AUC}     & \textbf{FENCE}                 \\ 
\midrule
\textbf{}                                                                                                            & \textbf{Clean} & 97.54                & 0.00                           \\
\textbf{Neris (1,9)}                                                                                                 & \textbf{PGD}   & 97.30                & 100                            \\
\textbf{}                                                                                                            & \textbf{FENCE} & 96.61                & 100                            \\ 
\midrule
\textbf{}                                                                                                            & \textbf{Clean} & 63.52                & 84.85                          \\
\textbf{Rbot (10,11)}                                                                                                & \textbf{PGD}   & 63.52                & 100                            \\
\textbf{}                                                                                                            & \textbf{FENCE} & 59.64                & 100                            \\ 
\midrule
\textbf{}                                                                                                            & \textbf{Clean} & 79.96                & 70.84                          \\
\textbf{Virut (13)}                                                                                                  & \textbf{PGD}   & 78.75                & 100                            \\
\textbf{}                                                                                                            & \textbf{FENCE} & 71.91                & 100                            \\
\bottomrule
\end{tabular}
\end{table}

\subsection{{Results -- RQ2 (Budget)}}

This research question is not relevant in the botnet case. Under the same budget of epochs and attack iterations, PGD-adversarial training already outperforms FENCE-adversarial training by 0.7\% in the previous research question. There is no significant utility in increasing the budget of examples to generate.   

\subsection{{Results -- RQ3 (Properties)}}

Our RQ1 results demonstrate that unrealistic hardening (PGD) and realistic hardening (FENCE) yield similar robustness against problem space attacks. Based on our findings in the text classification case, we hypothesize that both attacks explore the same search space, and that real botnets lie close to this explored vulnerable space.

\paragraph{\textit{Embeddings of the adversarial examples}}

\begin{figure}[ht!]
\centering
  \begin{subfigure}{\linewidth}
  \includegraphics[width=\linewidth]{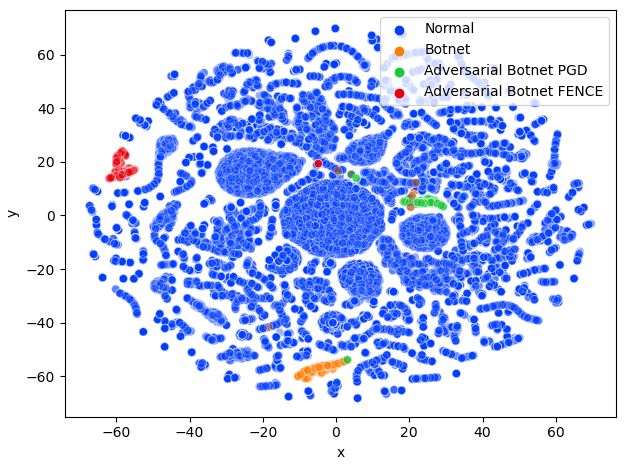}
  \end{subfigure}\hfill
  \caption{t-SNE visualization of original  and  successful adversarial examples generated by PGD and FENCE.}
  \label{fig-botnet-rq3}
  \end{figure}

In Figure \ref{fig-botnet-rq3} we visualize the clean and adversarial examples that each attack generated using t-SNE on Neris dataset. We see that both PGD and FENCE generate adversarial examples targeting particular neighbourhoods, forming ``islands" of adversarial examples. Varying the value of the perplexity  and random state  did not  impact majorly the general structures formed by the normal traffic examples. However, they  impacted the position of these ``islands". Fence and PGD examples were always  forming independent clusters from each other, however their position in the plot was not stable. They were sometimes closer and sometimes further from each other, and initial botnet examples while preserving their structure. This indicates that PGD and FENCE direct their examples to single blind-spots, however from t-SNE visualization itself we can not conclude that these spots are in close neighbourhoods.


\paragraph{\textit{Perturbation size and direction.}} Table \ref{cosine-botnet} presents cosine similarity and aggressiveness values for adversarial examples. The direction of PGD and FENCE adversarial examples is almost  the same, with their mean cosine similarity very close to 1 for the three datasets, respectively 0.88 for Neris, 0.97 for Rbot and 0.90 for Virut. Moreover, they are similarly aggressive because the difference between their aggressiveness scores is low. Respectively 0.37 for Neris, 0.03 for Rbot and 0.41 for Virut.  We conclude that this is one of the reasons why PGD is able to protect against FENCE examples.

\begin{table}[ht!]
\caption{Mean and standard deviation values for cosine similarity and aggressiveness of adversarial examples. }
\label{cosine-botnet}
\begin{tabular}{@{}llllll@{}}
\toprule
\textbf{Dataset} & \textbf{Attack} & \multicolumn{2}{l}{\textbf{Cosine similarity}}   & \multicolumn{2}{l}{\textbf{Aggressiveness}} \\
                 &                 & \multicolumn{2}{l}{\textbf{with FENCE examples}} & \multicolumn{2}{l}{}                        \\ \midrule
                 &                 & Mean         & \multicolumn{1}{l|}{Std}          & Mean                 & Std                  \\
Neris            & PGD             & 0.88         & \multicolumn{1}{l|}{0.05}         & 0.12                 & 0.05                 \\
                 & FENCE           & 1            & \multicolumn{1}{l|}{0}            & 0.49                 & 0.22                 \\ \midrule
Rbot             & PGD             & 0.97         & \multicolumn{1}{l|}{0.02}         & 0.06                 & 0.01                 \\
                 & FENCE           & 1            & \multicolumn{1}{l|}{0}            & 0.09                 & 0.13                 \\ \midrule
Virut            & PGD             & 0.90         & \multicolumn{1}{l|}{0.09}         & 0.13                 & 0.06                 \\
                 & FENCE           & 1            & \multicolumn{1}{l|}{0}            & 0.54                 & 0.43                 \\ \bottomrule
\end{tabular}
\end{table}

\begin{figure*}
  \begin{subfigure}{0.33\linewidth}
  \includegraphics[width=\linewidth]{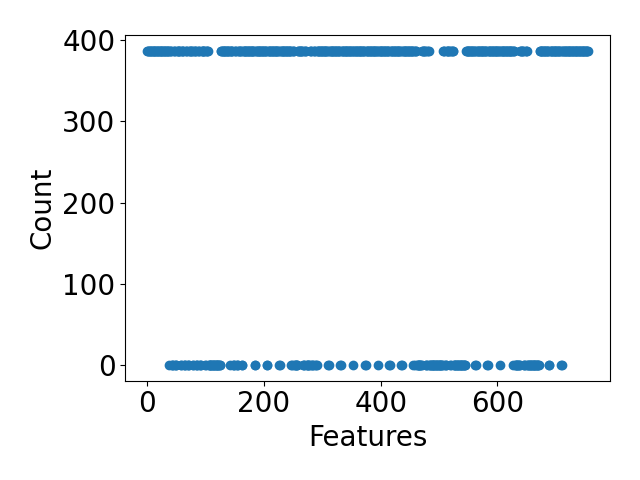}
    \caption{PGD}
  \end{subfigure}\hfill
  \begin{subfigure}{0.33\linewidth}
  \includegraphics[width=\linewidth]{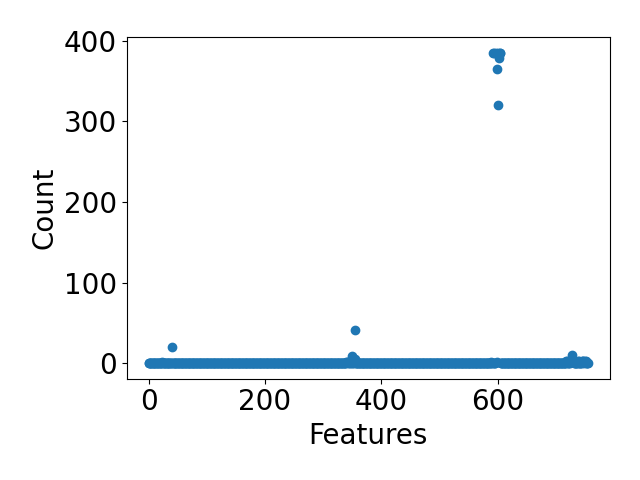}
    \caption{FENCE}
  \end{subfigure}\hfill
    \begin{subfigure}{0.33\linewidth}
  \includegraphics[width=\linewidth]{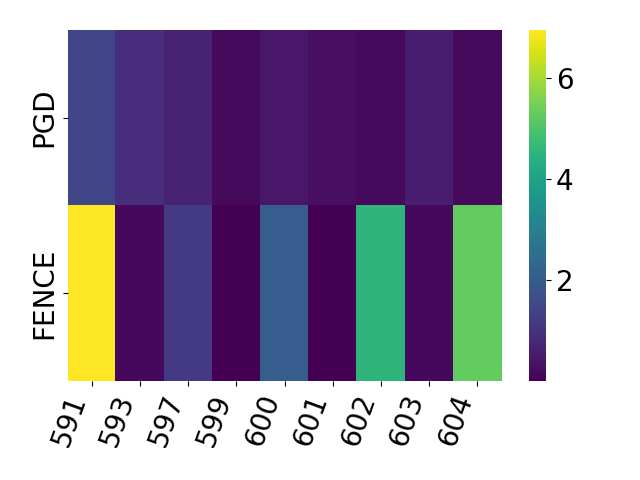}
    \caption{Average perturbation size }
  \end{subfigure}
  \caption{Analyses of features perturbed by PGD and FENCE. Subfigures a and b show, for each feature, the number of examples where the feature has been perturbed by PGD and FENCE, respectively. Subfigure c shows the average perturbation size caused by PGD and FENCE to the features that are perturbed more than 200 times by FENCE.}
  \label{fig-botnet-rq3-features}
  \end{figure*}

As a second step, we investigate whether PGD and FENCE perturb the same features.  Figures \ref{fig-botnet-rq3-features}a and \ref{fig-botnet-rq3-features}b show the number of examples where each feature has been perturbed by PGD and FENCE, respectively. We observe that PGD perturbs all the features that are mutable for all botnet samples. Meanwhile, FENCE has a small set of features that are perturbed in more than half of the samples, but most of them are never updated. To show the magnitude of the perturbation, in Figure \ref{fig-botnet-rq3-features}c, we display the average perturbation size for features that were updated in more than 200 samples by FENCE. The heatmap shows that PGD perturbs all features almost equally and with low effect, except for feature 591 -- the most perturbed. Feature 591 is also the feature that FENCE perturbs the most, and one of the only two that FENCE significantly perturbs. This means that perturbing feature 591 is the main factor to make an example reach the non-botnet class, and the two attacks exploit this feature to generate their adversarial examples. To confirm, we check the feature 591, and it is the feature related to the bytes sent from port 443 named \textit{bytes\_out\_sum\_s\_443}. This port is indeed commonly exploited by botnet traffic. Hence, its values are exploited by the attacks to pass a botnet example as normal traffic.

Ultimately, we explain the success of PGD in protecting against FENCE by two facts: (1) both PGD and FENCE examples follow the same direction and are similarly aggressive, (2) both FENCE and PGD apply the highest perturbation to the same feature.


%

\textbf{}

\subsection{{Conclusion for the Botnet Detection Use Case}}

In this section, we studied the hardening of the models that detect botnet traffic based on network flows. For this use case, an adversary is interested to ``disguise" botnet traffic  as normal traffic while preserving some inherited domain constraints that are related to packet size, time, protocol and ports relationship, etc. An attack that respects these constraints is the gradient based attack FENCE. We study how two different strategies of adversarial training affect the models' robustness against FENCE attack. The first strategy includes in the training process adversarial examples generated by the unconstrained attack PGD, and the second includes realistic constrained examples generated by FENCE. The results showed that despite the fact that PGD does not generate realistic examples, it was able to increase the robust accuracy of the model by 96\%, similar to 95.30\% by FENCE itself. The reason for this increase in robustness seems to be that the examples generated from PGD and FENCE
follow the same direction and are similarly aggressive. In addition, they have similar characteristics in feature perturbation. Being able to use PGD adversarial training to defend against a constrained attack like FENCE, brings benefits to the training costs, because PGD runs much faster than FENCE. 

\section{Windows Malware Detection}
\label{sec:malware}


Malicious software has always been a threat to the security of IT systems. With the recent surge of machine learning, research and practice have developed many learning-based detection systems that are able to distinguish between benign and malicious software \cite{ucci2019survey}. However, the same systems are also vulnerable to adversarial attacks that syntactically modify the malware without reducing its malicious effect \cite{castro2019aimed,pierazzi2020intriguing}. These attacks typically work in the problem space, e.g. they modify the malware PE file. Pierazzi et al. \cite{pierazzi2020intriguing} expose the conditions for a problem-space attack to be realistic, including preserved semantics, plausibility, robustness to preprocessing. We consider one such technique -- AIMED \cite{castro2019aimed} -- as the realistic attack to protect against. For model hardening, we consider different unconstrained feature space attacks, as well as constrained feature space attacks that capture some properties and dependencies of the features but cannot guarantee that the altered features can be mapped back to a real malware.

\subsection{Experimental Protocol}

\textbf{Realistic adversarial attack (to protect against).} We consider AIMED, a problem-space black-box attack \cite{castro2019aimed}. AIMED injects perturbations to the malware PE file in order to create variants of the malware, and runs a genetic algorithm to make these variants evolve. The attack aims to optimize four components: functionality, misclassification, similarity, and generation number. As recommended  by the results of the original AIMED implementation \cite{castro2019aimed}, the number of perturbations is 10 in order to reduce invalid examples. The generated malware files are  sent  for execution to a sandbox, for a functionality check. The sandbox setup is the one used by \cite{castro2019aimed} for AIMED evaluation. If functional, statistical features are extracted and fed to the classifier, which evaluates if the perturbed malware is evasive. On average, it takes 1 - 2 minutes to process a single example of PE files.

\textbf{Unrealistic adversarial attacks (for hardening).} We again use the PGD implementation from ART \cite{nicolae2018adversarial} as an unconstrained unrealistic attack. We also consider MoEvA2 \cite{simonetto2021unified}, a generic framework for constrained feature-space adversarial attacks that are based on genetic algorithms. MoEvA2 comes with a generic objective function that considers perturbation size, classification results, and constraint satisfaction. We identify immutability constraints based on the PE format description and structure given by Microsoft. We also extract feature dependency constraints from the original PE file examples we collected and those generated by AIMED. For example, the sum of binary features set to 1 that describe API imports should be less than the value of features \texttt{api\_nb}, which represents the total number of imports on the PE file. In our experiments, we use MoEvA2 as described above, as well as an alternative version where we remove the constraints in order to simulate an unconstrained attack. We will refer to this version as \emph{U-MoEvA2}.

As a preliminary check, similarly to the botnet use case, we have evaluated the attacks’ level of realism using the  analytical feature constraints derived from Simonetto et al.  for malware use case. As a reminder, respecting these constraints is a necessary but not sufficient condition for having realistic examples, since the problem space is more restricted than the feature space. Failure to satisfy  the contstraints indicates unrealistic examples. For this use case, respectively 0\% of PGD, 2\% of U-MoEvA2, 100\% of MoEvA2 and 100\% of AIMED adversarial examples respect the constraints. For U-MoEvA2 and MoEvA2 there is no guarantee that we can project back to the problem space since the problem -feature space relation is not invertible. On the other hand, AIMED generates an actual problem space object (Windows PE file) that are already checked for functionality. This confirms that PGD, U-MoEvA2 and MoEvA2  attacks are unrealistic in malware scenario, meanwhile AIMED is guaranteed to be realistic.




\textbf{Dataset and Model.} We use a collection of benign and malware PE files provided in \cite{aghakhani2020malware} to train a Random Forest model. You can find more details about the dataset and model in Appendix \ref{app:malware_exp}.

\textbf{Hardening strategy.}
To harden the model, we adversarially retrain the model with examples generated by PGD, U-MoEvA2, MoEvA2, and AIMED and retrain the classifier. Due to the costs related to generating adversarial examples with AIMED, we randomly select a sample of 1500 original malware examples, and we use only these examples to generate adversarial examples for retraining with the four attacks.

For adversarial examples generation, we study different ranges of $\epsilon$ thresholds. PGD does not change the maximum $L_2$ distance of its generated adversarial examples after $\epsilon = 0.1$. Additionally, the maximum $L_2$ distance at the threshold of 0.1 is much lower than the maximum $L_2$ achieved by AIMED. A threshold of 0.1 is also enough for MoEvA2 to generate adversarial examples that were later used for retraining. 

To generate adversarial examples with PGD, we train on the original training set a neural network with dense layers that achieves a ROC AUC score of 0.91, equal to the Random Forest. For the number of PGD iterations, we keep the default value in ART (100). 

\textbf{Evaluation.}
For evaluation, we use the subset of test examples where the prediction of the model and the actual label corresponds to malware (1069 for the clean model). We do not evaluate the robustness of the normal class because we consider that an attacker has no interest in our knowledge into fooling a classifier to identify a normal file as malware. 


\subsection{Results -- RQ1 (Hardening)}

Hardening the model with unrealistic examples does not improve the robustness of malware detection models against realistic adversarial attacks, even when the unrealistic examples respect some validity constraints.
Table \ref{tab-malware-rq1} shows for each model the clean accuracy and the robust accuracy against the realistic adversarial examples generated by AIMED. The clean model has an accuracy of 76.25\% on AIMED examples. The model benefits from retraining only with examples generated by AIMED itself (+18.99\%). Retraining with PGD gives a negligible improvement of  0.58\% and retraining with MoEvA2 or U-MoEvA2  slightly decreases the robust accuracy. In addition, hardening with AIMED does not hurt clean performance -- it actually improves it by 1.60\%.


\begin{table}
\centering
\caption{Clean and robust accuracy (against AIMED applied to the malware class) of the clean model  and the adversarially retrained models.}
\label{tab-malware-rq1}
\begin{tabular}{l|ll} 
\toprule
                &       & Attacked by  \\ 
\hline
Trained against & -     & AIMED        \\ 
\hline
Clean           & 90.60 & 76.25        \\
PGD             & 90.60 & 76.83        \\
MoEvA2          & 90.60 & 75.75        \\
U-MoEvA2        & 90.71 & 75.35        \\
AIMED           & 91.20 & 95.24        \\
\bottomrule
\end{tabular}
\end{table}



\subsection{Results -- RQ2 (Budget)}

\begin{figure}[ht!]
\centering
  \begin{subfigure}{\linewidth}
  \includegraphics[width=\linewidth]{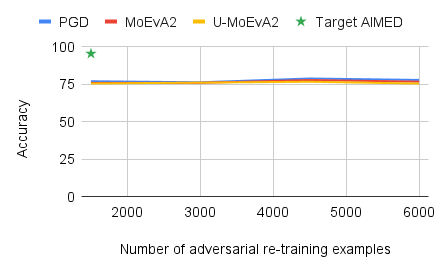}
  \end{subfigure}\hfill
  \caption{Robust accuracy against AIMED when increasing the number of adversarial retraining examples from 1500 to 6000, using each unrealistic attack. The green star represents the robust accuracy achieved retraining with 1500 adversarial examples generated by AIMED. }
  \label{fig-malware-rq2}
  \end{figure}

We, next, increase the number of adversarial examples generated by unrealistic attacks from 1,500 to 6,000 (using different original examples) and we observe the resulting robust accuracy against AIMED. Adversarial retraining with unrealistic examples does not benefit from this increased budget. 
Figure \ref{fig-malware-rq2} shows that the robust accuracy of PGD, U-MoEvA2, and MoEvA2  is constant against realistic adversarial examples generated by AIMED.



\subsection{Results -- RQ3 (Properties)}

Our results demonstrate that feature space hardening remains inferior to problem space hardening even with additional budget. We hypothesize that problem space attacks and feature space attacks (even with constraints) do not explore the same search space, and that problem space attacks better cover the vulnerable space of malware detection models.

\paragraph{\textit{Embeddings of the adversarial examples}}

\begin{figure}[ht!]
\centering
  \begin{subfigure}{\linewidth}
  \includegraphics[width=\linewidth]{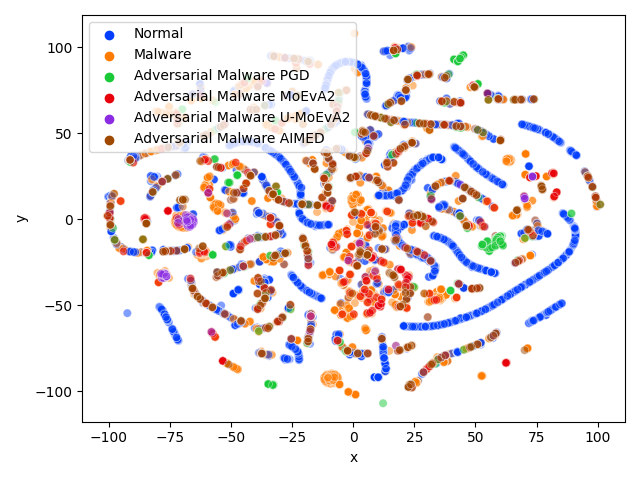}
  \end{subfigure}\hfill
  \caption{t-SNE visualization of original examples and the successful adversarial examples generated by PGD, U-MoEvA2, MoEvA2 and AIMED.}
  \label{fig-malware-rq3-embedings}
  \end{figure}

In Figure \ref{fig-malware-rq3-embedings} we visualize, using t-SNE, the embeddings of the clean training set and of the adversarial examples that each attack has generated.
Varying the value of the perplexity and random state did not impact the major structures formed in the graph. The hyperparameters only impacted the orientation and localization of the clusters.
We see that PGD examples form 2 major clusters. Similarly, examples from U-MoEvA2 form two clusters in two different neighbourhoods. Meanwhile, MoEvA2 examples do not particularly fall on the same neighbourhood. They are more spread throughout the graph.  Lastly,  the examples generated by AIMED cover a larger surface of the graph, with most of them following the embedding patterns of the normal files.


\paragraph{\textit{Perturbation size and direction}} Table \ref{cosine-malware} presents  cosine similarity and aggressiveness values of adversarial examples. The adversarial  examples generated by PGD, U-MoEvA2 and MoEvA2 follow different directions compared to AIMED examples. The mean cosine similarity to AIMED examples is equalling 0.39 for all of them. As a reminder, a cosine similarity of 0 indicates orthogonal vectors. Regarding the aggressiveness, AIMED adversarial examples are clearly more aggressive than unrealistic adversarial examples. They have a difference in mean aggressiveness value of 42.24 with PGD and MoEvA2 and 42.25 with U-MoEvA2. 

These observations are clearly different from text classification and botnet detection use cases, where  adversarial examples are more cosine-similar  and have the same level of aggressiveness. As a result, we conclude that this explains why hardening with unrealistic examples does not robustify the model against realistic examples in this use case.

\begin{table}[ht!]
\caption{Mean and standard deviation values cosine similarity and aggressiveness of adversarial examples. }
\label{cosine-malware}
\begin{tabular}{lllll}
\hline
\textbf{Attack}               & \multicolumn{2}{l}{\textbf{Cosine similarity}}   & \multicolumn{2}{l}{\textbf{Aggressiveness}} \\
                              & \multicolumn{2}{l}{\textbf{with AIMED examples}} & \multicolumn{2}{l}{}                        \\ \hline
\multicolumn{1}{l|}{}         & Mean         & \multicolumn{1}{l|}{Std}          & Mean                 & Std                  \\
\multicolumn{1}{l|}{PGD}      & 0.39         & \multicolumn{1}{l|}{0.17}         & 0.02                 & 0.04                 \\
\multicolumn{1}{l|}{U-MoEvA2} & 0.39         & \multicolumn{1}{l|}{0.17}         & 0.01                 & 0.07                 \\
\multicolumn{1}{l|}{MoEvA2}   & 0.39         & \multicolumn{1}{l|}{0.17}         & 0.02                 & 629.34               \\
\multicolumn{1}{l|}{AIMED}    & 1            & \multicolumn{1}{l|}{0}            & 42.26                & 0                    \\ \hline
\end{tabular}
\end{table}


\begin{figure*}[ht!]
  \begin{subfigure}{0.25\linewidth}
  \includegraphics[width=\linewidth]{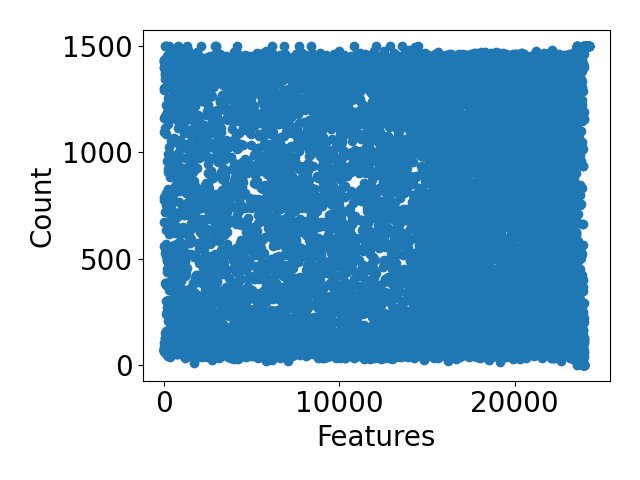}
    \caption{PGD}
  \end{subfigure}\hfill
  \begin{subfigure}{0.25\linewidth}
  \includegraphics[width=\linewidth]{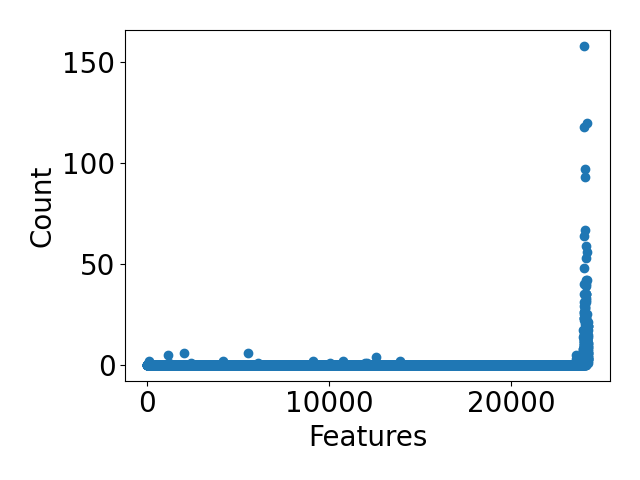}
    \caption{U-MoEvA2}
  \end{subfigure}\hfill
    \begin{subfigure}{0.25\linewidth}
  \includegraphics[width=\linewidth]{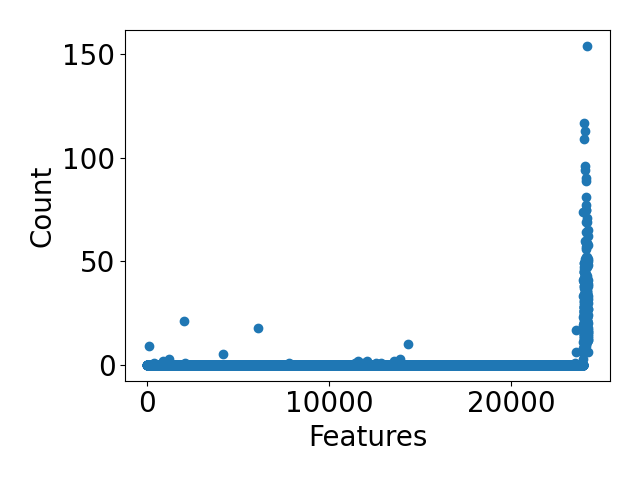}
    \caption{MoEvA2}
  \end{subfigure}\hfill
    \begin{subfigure}{0.25\linewidth}
  \includegraphics[width=\linewidth]{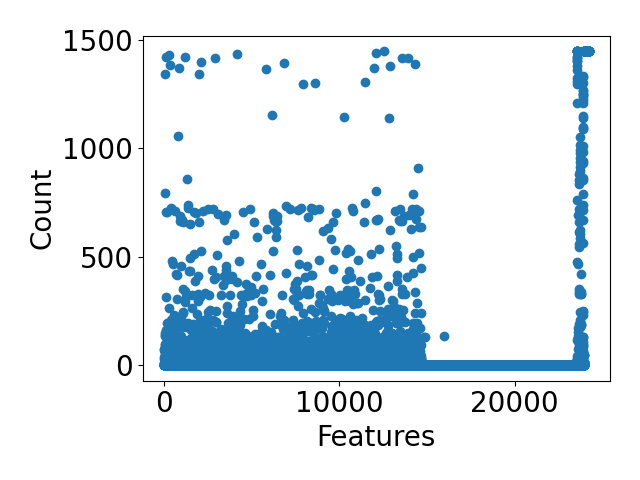}
    \caption{AIMED}
  \end{subfigure}
  \caption{Number of times each feature is perturbed in the generated adversarial examples used for retraining by PGD, U-MoEvA2, MoEvA2 and AIMED. }
  \label{fig-malware-rq3}
  \end{figure*}

In Figure \ref{fig-malware-rq3} we show, for any feature $f$, the 
number of adversarial examples that have a different value of $f$ compared to the original example it was generated from. PGD (\ref{fig-malware-rq3} (a)) perturbs almost all features. This is an expected behaviour as only 141 features out of 24 222 are not modifiable. The modified features are updated by PGD based on the direction of the gradient. We see in \ref{fig-malware-rq3} (b) and (c) that both U-MoEvA2 and MoEvA2 perturb similarly only the last features that are related to the byte frequency. It seems that the constraints related to these features are easier to satisfy, so MoEvA2 does not explore the full search space.

In \ref{fig-malware-rq3} (d) we see that AIMED has a different perturbation pattern compared to the three previous attacks. Similarly to MoEvA2, it perturbs at the end features related  to byte frequencies but as well other features in front of them. Another difference among them is that AIMED perturbs  the first half of the features, which are related mostly to API imports. 350 features are perturbed more than in half examples, with 266 of them perturbed always. AIMED avoids perturbing the middle features related to $dll$ imports, although these features are mutable. This is because the domain-specific transformations that AIMED implements never import $dll$.



\subsection{Conclusion for the Malware Use Case}

Our evaluation shows that for the malware classification use case, feature space hardening using PGD and U-MoEvA2, even when driven by domain constraints (MoEvA2) fails to harden the model. We demonstrate that even when given 4 times more budget, the absence of noticeable effect remains. Our investigation reveals that the realistic attack produces adversarial examples that follow different directions from the unrealistic examples. Moreover, the realistic attack is clearly more aggressive than the unrealistic attacks. The unrealistic feature-space attacks tend to modify either a small set of features or almost all features, differently from problem space attack AIMED. AIMED is capable of generating adversarial examples similar to the clean examples, which feature-space attacks are incapable of doing.  Consequently, these areas of the model's decision boundaries are not explored by feature space attacks and remain vulnerable to the realistic attack.


\section{Wrap-Up and Discussion}

Across our three use cases, we have observed that the effectiveness of adversarial hardening using unrealistic examples significantly varies: it ranges from not increasing model robustness at all (malware case) to achieving 99.70\% of robust accuracy (botnet case), or sometimes offering partial protection (text case). These results suggest the absence of a general trade-off between the realism of the produced examples and the potential of using these examples to protect against real-world attacks. This maintains hope that research on traditional adversarial attacks can still be useful to protect against domain-specific real-world attacks, and these domains go beyond computer vision. This is important to find a relative balance between the effort of a malicious third party to attack the system (i.e. design the domain-specific realistic attack and run it) and the effort of engineers to defend the system (i.e. through adversarial hardening). For practitioners and researchers, this suggests that \textit{unrealistic examples may help under strict conditions, hence they are worth a try}.

In all our use cases, we have observed that increasing the number of generated unrealistic examples has negligible effects on robustness. This suggests that there is an asymptotic limit to the utility of unrealistic examples in model hardening and defending against realistic attacks effectively is not a matter of computational budget involved, but is rather a matter of the characteristics of these generated examples. Therefore, improving the efficiency of unrealistic attacks — a dedicated area of research — is worthless in these settings. Research should rather work on profoundly changing these attacks. Our takeaway is that \textit{if unrealistic examples do not bring improvement even at a small scale, they  will never do,  regardless of the allocated computational budget.}

Through our in-depth analysis of all adversarial examples, we have highlighted that unrealistic examples are useful only if they have a close representation to the realistic examples (i.e. they follow the same direction, they are similarly aggressive, and  they perturb the same features). Indeed, the botnet case has revealed that unrealistic examples and realistic examples overlap in the feature space, and the attacks that generated these examples typically modify the same features. \textit{Our results pave the way for developing new adversarial hardening mechanisms. These mechanisms would rely on new attacks (or adaptation of existing attacks) to efficiently produce unrealistic examples that mimic realistic adversarial examples.} This  follows the assumption that a set of realistic adversarial examples are available (e.g. captured from previous attack attempts or produced by an existing attack). As a future work, one of those considered mechanisms is learning domain constraints from realistic examples, then integrating the constraints as an attack objective (e.g. into PGD). Another foreseen mechanism would drive the attack in a direction towards realistic examples — achieving a high cosine similarity and a similar degree of aggressiveness — in order to produce examples that are close in their latent space.

Our replication package (including the datasets and attacks we used) can act as a benchmark for future research to lean on. In particular, the malware use case has required considerable engineering effort and significant computation resources to produce the examples and harden the models.

\section{Conclusion}

Since the seminal work that uncovered the vulnerability of machine learning models to adversarial examples\cite{biggio2013evasion}, researchers and practitioners have realized the potential harm that adversarial examples could cause in the real-world \cite{pierazzi2020intriguing, schonherr2020imperio}. Realistic adversarial attacks remain more challenging to design and, therefore, harder to study and protect against. To our knowledge, this work is the first evaluation of the limitations and opportunities of adversarial hardening against realistic adversarial examples, and it is conducted across multiple domains. We analyze three real-world use cases: {text classification}, {botnet detection} and {malware classification} and demonstrate that cheap but unrealistic adversarial examples can, under specific conditions, successfully harden models against realistic but expensive attacks. We reveal these conditions to explain when unrealistic adversarial examples are relevant to harden against realistic attacks, and provide insights to practitioners to help them decide which strategy to adopt to protect their models against real-world attacks.

\section*{Acknowledgments}
Salijona Dyrmishi's work is supported by the Luxembourg National Research Funds(FNR) AFR Grant 14585105.

\bibliographystyle{IEEEtran}
\bibliography{references}
\newpage
\clearpage
\appendices

\section{{Related work}}
\label{sec:related}

\subsection{{Constrained adversarial attacks in feature space}}
Most of the constrained  attacks in feature space are an upgrade of traditional computer vision attacks like JSMA \cite{papernot2016limitations}, C\&W \cite{carlini2017towards} or FGSM \cite{goodfellow2014explaining}. The majority of these attacks are evaluated in Network Intrusion Detection Systems (NIDS) \cite{chernikova2019fence, sheatsley2020adversarial, teuffenbach2020subverting, tian2020exploring, sheatsley2021robustness} and few other domains: Cyber-Physical Systems \cite{li2020conaml}, Twitter bot detection and website fingerprinting \cite{kulynych2018evading}, and credit scoring \cite{ghamizi2020search, levy2020not, cartella2021adversarial}. Every attack handles different types of constraints,  resulting in loosely realistic to totally realistic adversarial examples.

Kulynych et al. \cite{kulynych2018evading} proposes a graphical framework where the nodes represent the feasible examples and the edges represent valid transformations for each feature. 
Despite the results, this A* based attack is practically infeasible in many large datasets due to its time complexity.
Levy et al. \cite{levy2020not} updates JSMA attack to take into account immutable features and feature distribution. 
Cartella et al. \cite{cartella2021adversarial} proposes some modifications to three existing attacks: Zoo, Boundary attack, HopSkipJump. The modifications consider the following factors: feature modifiability, feature data types, and feature importance. All these three attacks \cite{kulynych2018evading, levy2020not, cartella2021adversarial} mentioned above consider only constraints on individual features and neglect inter-feature relationships, therefore there is a high risk that attacks generate totally unrealistic examples. Another attack which considers only individual features was proposed by Teuffenbach et al. \cite{teuffenbach2020subverting}.  They divided the features in groups based on the difficulty to attack them. They extend C\&W attack to consider weights in these groups and favor modifications on low weighted features, i.e. the independent features. This approach does not guarantee that the generated adversarial examples respect constraints. Instead, it hopes that only modifications in independent features will be enough to cause a wrong prediction.

Among the attacks that consider feature relationships are the ones introduced in \cite{tian2020exploring}, \cite{li2020conaml}, \cite{sheatsley2020adversarial}, \cite{sheatsley2021robustness}, \cite{chernikova2019fence}, \cite{ghamizi2020search}, \cite{simonetto2021unified}.   Tian et al. \cite{tian2020exploring}  update IFGSM so that the size and direction of the perturbation are restricted to respect feature correlations. Despite this method being among the few that use a large variety of constraints types, it can not handle relationships of more than two features. As a result, their constrained IFGSM attack generates realistic examples only in domains where there are no relationships between more than two features.  Li et al. \cite{li2020conaml} identify a series of linear constraints  related to sensor measurements  for two  Cyber-Physical Systems domains and craft a  best-effort search attack that would respect these constraints. Their approach handles non-linear constraints and a limited type of simple non-linear relationships.

Sheatsley et al.  \cite{sheatsley2020adversarial} integrate constraint resolution into  JSMA  attack to limit feature values in the range dictated by their primary feature. Later on, for the same case study, Sheatsley et al. \cite{sheatsley2021robustness}, learned boolean constraints with Valiant algorithm and introduced Constrained Saliency Projection (CSP), an improved iterative version of their previous attack. The bottleneck of this attack is the process of learning the constraints. Its runtime  scales exponentially with the number of features. Chernikova et al. \cite{chernikova2019fence} designed a general attack framework named  FENCE for constrained application domains. The framework includes an iterative gradient attack that updates features based on their family dependencies. FENCE is one of the attacks that handles the largest type of constraints, resulting in realistic constrained feature space examples.  Ghamizi et al. \cite{ghamizi2020search} propose a genetic algorithm that encodes constraints as objectives to be optimized. They support a variety of linear, non-linear, and statistical constraints as long as they can be expressed as a minimization problem. This ensures that the generated constrained examples are realistic. Later, Simonetto et al. \cite{simonetto2021unified} followed a similar approach as \cite{ghamizi2020search} to propose a generic framework for generating constrained adversarial examples.

You can find a complete list of constrained adversarial attacks and their properties in Table~\ref{tab-cons-attacks}.


\begin{table*}[ht!]
\centering
\caption{Prior work on feature space adversarial attacks  that consider domain constraints}
\resizebox{\textwidth}{!}{%
\begin{tabular}{llllllll} 
\toprule
                                                  & Year & Use case                         & Threat model     & Method                                & Constraint type                 & Expansion to new domains    & Open source  \\ 
\midrule
\cite{kulynych2018evading}    & 2018 & Twitter bot                      & White \&  Black box & Graphical framework                   & Mutability                      & Define~                     & Yes          \\
                                                  &      & Website-fingerprinting           &                  &                                       & Range               & the transformation graph    &              \\ 
\hline
\cite{chernikova2019fence}       & 2019 & Botnet detection                 & White box        & Iterative Gradient Based               & Mutability, Range,              & Rewrite most~               & Yes          \\
                                                  &      & Malicious domain classification  &                  &                                       & Ratio,One-hot encoding          & of the components           &              \\
                                                  &      &                                  &                  &                                       & Statistical ,Linear, Non-linear &                             &              \\ 
\hline
\cite{ghamizi2020search}         & 2020 & Credit scoring                   & Grey-box         & Genetic Algorithm                     & Mutability, Range,              & Define constraints~         & Yes          \\
                                                  &      &                                  &                  &                                       & Ratio, One-hot-encoding,        & as a minimization problem   &              \\
                                                  &      &                                  &                  &                                       & statistical, linear, non-linear &                             &              \\ 
\hline
\cite{li2020conaml}              & 2020 & Water treatment attack detection & White \& Black box  & Best-Effort Search                    & Mutability,range                & Define constraints          & No           \\
                                                  &      & State Estimation in Power Grids  &                  &                                       & Linear, 
Simple non-linear      &                             &              \\ 
\hline
\cite{levy2020not}               & 2020 & Price of lodging (AirBnb)        & White-box        & Extended JSMA                         & Mutability, Range,              & Yes                         & No           \\
                                                  &      & Medical data ICU                 &                  &                                       &                                 &                             &              \\
                                                  &      & Credit risk                      &                  &                                       &                                 &                             &              \\ 
\hline
\cite{sheatsley2020adversarial}  & 2020 & NIDS                             & White-box        & Constrained Augmented JSMA            & Mutability,Range,               & Learn constraints~          & No           \\
                                                  &      &                                  &                  &                                       & Boolean                         & from data                   &              \\ 
\hline
\cite{teuffenbach2020subverting} & 2020 & NIDS                             & White-box        & Extended CW attack                    & Mutability                      & Define feature groups       & No           \\ 
\hline
\cite{tian2020exploring}         & 2020 & NIDS                             & White-box        & Constraint-based IFGSM                & Linear                          & Define constraints~         & No           \\
                                                  &      &                                  &                  &                                       & Addition boundary constraint,   & by visual exploration       &              \\
                                                  &      &                                  &                  &                                       & Zero multiplication constraint  &                             &              \\ 
\hline
\cite{cartella2021adversarial}   & 2021 & Fraud                            & Black box        & Modified Zoo, Boundary attack, HopSkipJump         & Mutability,Data type            & Define feature weights      & No           \\ 
\hline
\cite{sheatsley2021robustness}   & 2021 & NIDS                          & White-box        & Constrained Saliency Projection & Boolean constraints             & Learn constraints~          & No           \\
                                                  &      & Phishing                         &                  &                                       &                                 & through Valiant’s algorithm &              \\
\bottomrule
\end{tabular}}
\label{tab-cons-attacks}
\end{table*}

\subsection{{Problem space attacks}}
Problem space attacks design and application are complex. Generally, it is easier to deal with digital objects than tangible ones, therefore there are more problem space attacks in the digital world. \cite{anderson2018learning} uses reinforcement learning to modify binary Windows PE files to evade  Windows malware detection engines. The generated adversarial files are not checked if they are valid, therefore some files can be corrupt. \cite{castro2019aimed} uses the same set of operations as \cite{anderson2018learning} combined with Genetic Algorithm to generate adversarial Windows PE files.  In the contrary to \cite{anderson2018learning}, Castro et al. evaluate in each iteration if the modified samples are still functional in order to avoid corrupt files. The generated files are valid and preserve their maliciousness. 
\cite{pierazzi2020intriguing}  introduces a new problem space attack for android malware detection. They initially collect a set of bytecodes (gadgets) from benign files through program slicing. At the attack stage, they select the most important features for the classifier and gadgets that include those features. The gadgets are then implanted on the malicious file.  By the design described above, the generated files are functional. 
\cite{schonherr2020imperio} introduces Imperio, an attack that produces adversarial audio examples for speech recognition systems. The particularity of the Imperio is that the audio can be played in different room environments without the need to be tuned for that specific environment. Additionally, the attack does not require a direct line-of-sight between speaker and microphone. Another problem space domain where there exists a large variety of adversarial attacks is text. There is a category of attacks like \cite{gao2018black, li2018textbugger, pruthi2019combating} that generate adversarial text by replacing/swapping characters in the most influential words. The drawback of this approach is that the generated sentences will not continue to be adversarial if they are sent to a grammar checker before going to a classifier as input. Other attacks instead of replacing characters, replace complete words with their synonymous \cite{jin2020bert, ren2019generating, li2020bert}. These attacks are less suspicious to human eyes, however, if the number of replaced words is not controlled properly the sentence might lose its original meaning.



\section{Problem formulation}
\subsection{{Threat model}}
\label{app:threat-model}
\textbf{Attacker's Objective:}
The adversary aims to fool the classification model in a way that it can change the classification result. For instance,  the adversary aims to make a malicious behavior (e.g. malware or botnet traffic) be labelled as benign.

\textbf{Attacker's Knowledge:}
We assume that the attacker knows the test distribution of the model and has access to specific examples that the model classifies correctly. For example, these examples could be malware or botnet traffic correctly detected by the model (the attacker will use these examples as a starting point for the attack). The realistic attack we consider can be black-box  or white-box. In black-box attacks, the attacker is not aware of the model's specific information (architecture, weights, train dataset, etc). In a white-box attack, the attacker has access to all the above information. 
 

\textbf{Attacker's Capability:}
We assume that the attacker has the computational capacity to produce some realistic adversarial examples that successfully fool the model. This means that the attacker has designed such a realistic attack for the particular application domain of interest (e.g. an attack that modifies malware PE files). Hence, starting from an original example that is correctly classified, the attacker can produce an adversarial example that is semantically similar to the original one, but is misclassified by the model. Though realistic adversarial attacks are expensive to run (in particular, when using those attacks for model hardening), it is generally enough to apply them a few times to produce a successful example. In the use cases we consider in our experiments, the attacker has to apply these attacks up to four times on average to produce a successful example.




\subsection{{Defender's Model}}
\label{app:defender-model}
\textbf{Defender's Objective:}
The defender aims to improve the model in order for it to classify correctly the adversarial examples that the attacker produces.

\textbf{Defender's Knowledge:}
The defender has full access to the model and to both the training and test set that were used for training and validation. 

\textbf{Defender's Capability:}
We assume that the defender has no access to the adversarial attack method used by the attacker and/or the defender does not have the computational capacity to produce the number of adversarial examples required for model hardening. Therefore, the defender cannot harden the model with realistic examples. We, however, assume that the defender has access to unrealistic adversarial attacks that have cheaper computational costs, and can produce a sufficient number of unrealistic adversarial examples to harden the model. We consider both gray-box unrealistic attacks (e.g. evolutionary with the logits' based fitness function) or white-box attacks (e.g. gradient based).

\section{Natural language constraints}
\label{app:text-constraints}
Morris et al. \cite{morris2020reevaluating} define four types of constraints for a successful and realistic adversarial example in natural language tasks: \textit{ semantics, grammaticality, overlap, } and \textit{non-suspicion}. Semantics ensures that the adversarial transformations do not change the meaning of the sentence. Grammaticality requires that the adversarial examples do not introduce any grammatical errors. This is essential for an attack to be effective, since examples with grammar errors can easily be detected by a grammar checker. For example, using a verb tense of the future for an event happening in the present. Overlap constraints restrict the similarity between original and generated texts at the character level. Finally, non-suspicion comprises context-specific constraints that ensure the adversarial text does not look suspicious. For example, replacing a word in a modern English text with a synonym from the English of Shakespeare in an academic paper would raise suspicions. 

\section{Experimental Protocol}
\subsection{Text classification}
\label{app:text_exp}
\textbf{Datasets.}For the evaluation we use three datasets: 1) Rotten Tomatoes movie reviews \cite{pang2005seeing}, 2) Offensive subset from TweetEval dataset \cite{barbieri2020tweeteval}, 3) AG News \cite{Zhang2015CharacterlevelCN}. For AG News, we use only 7.5k samples for hardening  and  1k samples for evaluation. Both Rotten Tomatoes and Tweet Offensive are binary sentiment analysis datasets, while AG News classifies news in four categories. A training sample in these datasets is an input that consists of one or more sentences. The datasets come with different characteristics of input lengths. The smallest dataset is Rotten Tomatoes with 18.49 words in average per input. The largest dataset is AG News with an average of 38.78 words per input. 

\textbf{{Model.}}
We use DistilBERT \cite{sanh2019distilbert} base model (uncased) available in Hugging Face, because it achieves a comparable performance with BERT, but it is lighter and faster to train. This general pre-trained language model can be fine-tuned in any task. To evaluate the initial robustness of the clean model, we perform normal training for 5 epochs. The model's clean performance is comparable to the state of the art for the dataset used in this study.

\subsection{Botnet detection}
\label{app:botnet_exp}

\textbf{{Datasets.}}
For this case study, we use the standard CTU-13 dataset. To avoid training biases, we choose from this dataset bot types with at least two scenarios, one for training and one for testing. This filtering resulted in three bot types, Neris, Rbot and Virut. The selected datasets are feature engineered based on the process from \cite{ongun2019designing}, the same process used by FENCE \cite{chernikova2019fence}.  The datasets are scaled using a Standard Scaler. Neris has 143K training examples (scenario 1,9) and 55K testing examples (scenario 2), with only 0.74\% examples labelled in the botnet class. Rbot has 59k training examples (scenario 10,11) and 49k testing examples, with only 0.25\% examples labelled as botnet. 
Virut has 146k training examples and 5k testing examples, with only 0.74\% labelled as botnet.

\subsection{Windows Malware detection}
\label{app:malware_exp}
\textbf{Datasets.}
We use a collection of benign and malware PE files provided in \cite{aghakhani2020malware}. 
Machine learning classifiers based on static analysis are more prone to detect packing as a sign of maliciousness due to biases in the training set. Aghakhani et al. \cite{aghakhani2020malware} showed that if the training set is built such that includes a mix of unpacked benign and packed benign in addition to malicious executables, it is less biased  towards detecting specific packing routines as a sign of maliciousness. Therefore, we select in total 4396 packed benign, 4396 unpacked benign, and 8792 malicious samples. We select 70\% of files to use for training, 15\% for testing, and 15\% for adversarial training evaluation. From all our 17584 samples, we extract a set of static features which include: \textit{PE headers}, \textit{PE sections}, \textit{DLL imports},  \textit{API imports}, \textit{Rich Header}, \textit{File generic}. In total there are 24 222 features in this dataset.

\textbf{Model.} To classify benign and malware software, we used a Random Forest model, a largely used classification algorithm for this task in the literature. We use 100 trees as estimators. This model reached 0.91 AUC-ROC, similar to what other papers reported \cite{anderson2018learning}. 

\section{Adversarial embeddings t-SNE plots}
\label{app:plots}
\begin{figure}[ht!]
\centering
    \begin{subfigure}{0.8\linewidth}
  \includegraphics[width=\linewidth]{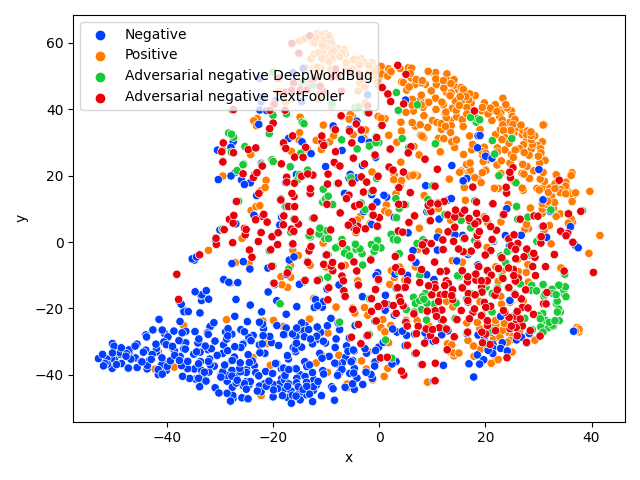}
    \caption{Rotten Tomatoes negative reviews}
  \end{subfigure}\hfill
  \begin{subfigure}{0.8\linewidth}
  \includegraphics[width=\linewidth]{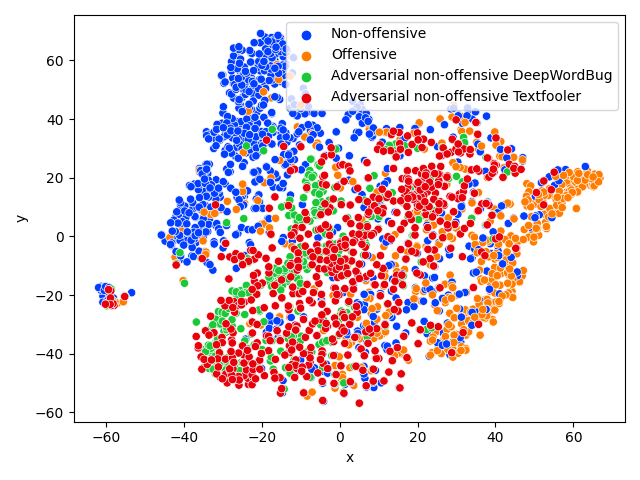}
    \caption{TweetEval (Offensive) non-offensive tweets}
  \end{subfigure}\hfill
  \begin{subfigure}{0.8\linewidth}
  \includegraphics[width=\linewidth]{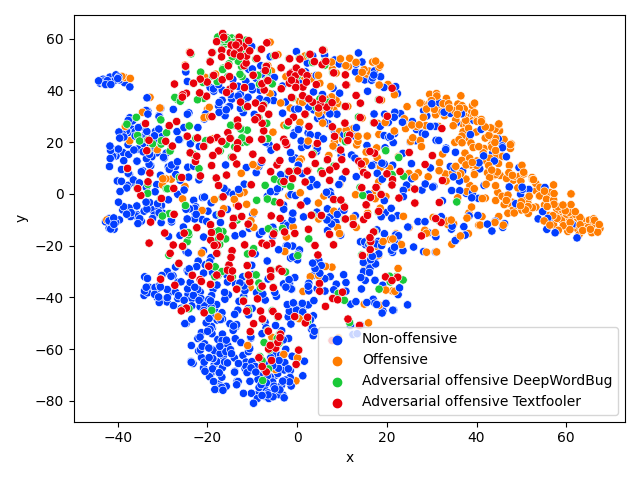}
    \caption{TweetEval (Offensive) offensive tweets}
  \end{subfigure}\hfill
  \caption{t-SNE sentence embeddings from the last layer latent representation of the models. The graphs include the original examples and the successful adversarial examples generated by DeepWordBug and TextFooler.}
  \label{fig-text-rq3-comp}
  \end{figure}
\end{document}